\pdfoutput=1

\documentclass[11pt]{article}

\usepackage[final]{acl}

\usepackage{times}
\usepackage{latexsym}
\usepackage{algorithmic}

\usepackage{graphicx}    
\usepackage[table,xcdraw]{xcolor}
\usepackage[T1]{fontenc}

\usepackage[utf8]{inputenc}

\usepackage{microtype}

\usepackage{inconsolata}

\usepackage[ruled]{algorithm2e}
\usepackage{graphicx}
\usepackage{multirow}
\usepackage{latexsym}
\usepackage{amsmath,amssymb,amsthm}
\usepackage{subfigure}
\usepackage{microtype}

\usepackage{booktabs,arydshln}
\usepackage{setspace}
\usepackage[normalem]{ulem}
\useunder{\uline}{\ul}{}
\usepackage{tikz}
\usepackage{bbding}
\usepackage{pifont}
\usepackage{wasysym}
\usepackage{paralist}
\usepackage{setspace}
\usepackage{tikz}
\usepackage{pgf-pie} 
\usepackage{subcaption}
\usepackage{xcolor}
\usepackage{float}

\title{What Gets Activated: Uncovering Domain and Driver Experts \\in MoE Language Models}

\author{Guimin Hu$^{1}$, Meng Li$^{2}$, Qiwei Peng$^{3}$, Lijie Hu$^{4}$, Boyan Xu$^{1}$, Ruichu Cai$^{1}$\\
  $^1$Guangdong University of Technology \\
  $^2$Soochow University \\
  $^3$University of Copenhagen \\
  $^4$Mohamed bin Zayed University of Artificial Intelligence \\
\texttt{rice.hu.x@gmail.com}}

\begin{document}

\maketitle
\begin{abstract}
Most interpretability work focuses on layer- or neuron-level mechanisms in Transformers, leaving expert-level behavior in MoE LLMs underexplored. Motivated by functional specialization in the human brain, we analyze expert activation by distinguishing domain and driver experts.
In this work, we study expert activation in MoE models across three public domains and address two key questions: (1) which experts are activated, and whether certain expert types exhibit consistent activation patterns; and (2) how tokens are associated with and trigger the activation of specific experts. To answer these questions, we introduce entropy-based and causal-effect metrics to assess whether an expert is strongly favored for a particular domain, and how strongly expert activation contributes causally to the model’s output, thus identify domain and driver experts, respectively. Furthermore, we explore how individual tokens are associated with the activation of specific experts. Our analysis reveals that (1) Among the activated experts, some show clear domain preferences, while others exert strong causal influence on model performance, underscoring their decisive roles. (2) tokens occurring earlier in a sentence are more likely to trigger the driver experts, and (3) adjusting the weights of domain and driver experts leads to significant performance gains across all three models and domains. These findings shed light on the internal mechanisms of MoE models and enhance their interpretability.
\end{abstract}

\section{Introduction}
Mixture-of-Experts (MoE) models have become increasingly prominent for their ability to scale neural architectures efficiently while achieving strong performance \cite{chen2024llava,yu2023mmoe,lin2024moe}. Their integration into large language models (LLMs) has led to impressive advancements in NLP \cite{radford2019language,brown2020language}, especially in Transformer-based LLMs by leveraging a gating function to dynamically select experts based on input data, enabling the model to handle complex, high-dimensional datasets \cite{radford2019language,brown2020language}. 


Prior mechanistic interpretability studies \cite{DBLP:conf/nips/MengBAB22,DBLP:conf/emnlp/HuoYHYH24,DBLP:conf/acl/TangLH0WZWW24} primarily focused on layer-level or neuron-level interpretations, while overlooking the operational mechanisms of MoE in LLMs. For example, \citet{yu2023mmoe} proposes multimodal mixtures of experts (MMOE) to train separate expert models for each type of multimodal interaction. \citet{gururangan2022demix} introduces Demix Layers, where different experts are specialized for distinct domains in NLP tasks, enabling domain-specific adaptation without fine-tuning the entire model. \citet{muennighoffolmoe} provides a thorough analysis of four MoE-specific phenomena: router saturation, expert co-activation, domain specialization, and vocabulary specialization.
However, most current MoE integration strategies adopt the framework as-is, leaving the internal mechanisms of MoE-based large language models largely unexplored. This lack of interpretability can lead to unpredictable or even harmful behaviors, raising concerns about the safety and reliability of their deployment \cite{hendrycks2023overview,ngo2022alignment}.

In the field of neuroscience, the brain is functionally specialized, exhibiting feature-selective, often sparse responses; a small subset of driver/hub neurons wields disproportionate causal influence, where minor interventions cause outsized changes \cite{sherman1998actions,kanwisher1997fusiform}. Motivated by such structure in the human brain, where regions such as Broca’s area and Wernicke’s area are responsible for specific types of information processing, we draw an analogy to the Mixture of Experts (MoE) architecture. 
We posit that MoE can be organized into distinct functional components—general experts that capture universal knowledge and specialized experts that handle specific tasks or mediate input–output causal effects. From this perspective, we introduce the notions of domain experts and driver experts into expert-activation analysis, aiming to reveal the underlying governing principles.

Our study focuses on MoE-based LLMs and investigates expert activation through two questions: (1) \textit{which experts are activated, and do certain types exhibit consistent activation patterns}; and (2) \textit{how are tokens associated with activated experts?} To this end, we develop entropy-based and causal-effect metrics to uncover activated specific experts. To identify \textbf{domain experts}, we introduce a domain-specific expert entropy that quantifies an expert’s specialization within a given domain. To identify \textbf{driver experts}, we measure causal effects on model outputs by perturbing routing gates and designate as drivers those experts that induce significant output changes. Using these entropy and causal-effect estimates, we quantify each expert’s contribution to predictions and analyze how adjusting the routing weights of domain and driver experts impacts performance.
\begin{itemize}
    \item We propose entropy- and causal-effect–based metrics to quantify an expert’s domain specialization and its causal influence on predictions, thereby enabling the identification of domain and driver experts.
   \item We analyze how specific input tokens trigger domain and driver experts, finding that earlier tokens in a sentence are more likely to activate them, suggesting opportunities for token-aware expert routing.
    \item On average across three MoE LLMs, tuning the weights of domain and driver experts yields accuracy gains of 2.08\% and 3.00\%, respectively, underscoring the practical value of task-aware expert activation in MoE models.
\end{itemize}

\section{Research Framework}

\subsection{Research Questions}

\paragraph{Expert Type Definition}
Inspired by feature-selective and driver neurons in the field of neuroscience \cite{sherman1998actions,kanwisher1997fusiform}, we identify activated experts and categorize them as general, domain-oriented, and driver experts.
\textbf{General expert} denotes an expert with no pronounced domain preference, whose routing/activation is approximately uniform across domains. \textbf{Domain expert} is not intended to imply strict exclusivity but rather relative preference to a target domain—preferentially routed on domain-specific inputs, with lower routing entropy and higher activation frequency in that domain.
\textbf{Driver expert} refers to experts with high causal influence in a target domain—perturbing it produces a significant causal effects in specific task performance. MoE gating tends to form a small set of “generalist experts” whose peaked activations naturally appear across tasks. Some of our datasets share linguistic or task structures (e.g., factual QA vs. open-domain reasoning), making it natural for overlap across domain and driver experts.

\paragraph{Which experts are activated?}
Language- and domain-specific neurons have been identified in \cite{DBLP:conf/acl/TangLH0WZWW24} and \cite{DBLP:conf/emnlp/HuoYHYH24}, respectively. A growing body of work investigates expert specialization, showing that different experts tend to specialize in particular tokens \cite{muennighoffolmoe}, domains \cite{jiang2024mixtral}, or tasks \cite{chaudhari2025moe}. These insights have enabled targeted optimization of specialized experts \cite{chen2022towards}. Inspired by these, our work investigates the expert specialization. This observation motivates the central question: \textit{Which experts are activated?}

\paragraph{How are tokens associated with specific experts?}
Prior work shows that stored information varies with token position \cite{DBLP:conf/nips/MengBAB22,DBLP:conf/iclr/ToddLSMWB24}, and recent results report that the first token in advanced language models disproportionately attracts attention \cite{barbero2025llms}. A gating network routes information by assigning each token (or segment) to a small subset of experts; accordingly, we study the token–expert mapping by pinpointing tokens that consistently elicit particular expert activations, raising the central question: \textit{How are tokens associated with specific experts?}



\subsection{Activated Expert Analysis}

\subsubsection{Domain Expert Identification}
Prior work \cite{chaudhari2025moe} show that MoE models tend to rely heavily on a small subset of highly specialized experts, with the top-weighted expert’s output often closely matching the full expert ensemble. Inspired by the work, we systematically investigates domain experts across three MoE language models.
A domain expert is identified by measuring the uncertainty of the gating network when selecting activated experts. The routing gate defines a probability distribution over experts conditioned on inputs from $M$ domains, denoted as $\{p(e_i|x_1), p(e_i|x_2), \dots, p(e_i|x_M)\}$ for expert $e_i$. To simplify the computation, we reduce the original $K$-way gating distribution to a binary distribution. Specifically: (1) selecting expert $i$ with probability $p_i$; and (2) selecting any expert other than $i$ with probability $1 - p_i$.
By grouping all non-$i$ experts into a single combined event, we obtain a convenient Bernoulli distribution that distinguishes between “select expert $i$’’ and “select some other expert.’’ This allows us to compute the activation entropy for expert $i$ within a specific domain (e.g., $D_{j}$) as follow:
\begin{align}
    &H_i(D_j) = \mathbb{E}_{x \sim D_j} \left[ -\tilde{p}(e_i \mid x) \log \tilde{p}(e_i \mid x) \right]\\
    &\tilde{p}(e_i \mid x) = 
\frac{\exp(z_i)}{\sum_{j \in \mathrm{Top}\text{-}k(x)} \exp(z_j)}
\end{align}
where $\tilde{p}(e_i \mid x)$ denotes the gating probability assigned to expert $e_i$ for an input $x \sim D_j$ under the Top-k routing scheme, which avoid the low entropy cause by “never used” expert. This metric reflects the specialization of $e_{i}$ on domain $D_{j}$. Under this scheme, the entropy of expert activation reveals that individual experts are activated with high certainty only for inputs drawn from certain data domains.

We measure whether an expert is activated primarily on specific data domains using the \emph{activation rate}. For expert $e_i$ and domain $D_j$, the activation rate is defined as
\begin{align}
    &A_i(D_j) = \mathbb{E}_{x \sim D_j}\big[ \mathbf{1}(e_i \text{ is activated}) \big]
\end{align}
A higher value of $A_i(D_j)$ on certain domains indicates that the expert is more frequently activated for those domains, while near-zero activation rates on other domains suggest clear domain preference.

To simultaneously quantify an expert’s activation certainty and its specialization on domain $D_{j}$, we introduce the Certainty-Weighted Activation Score (CWAS), defined as:
\[
S_i(D_j) = \big(1 - H_i(D_j)\big) \cdot A_i(D_j),
\]
where $A_i(D_j)$ quantifies how frequently expert $e_i$ is activated on domain $D_j$, and $1 - H_i(D_j)$ reflects the certainty of activation (lower entropy implies higher certainty). A high value of $S_i(D_j)$ indicates that expert $e_i$ is both frequently activated on domain $D_j$ (high $A_i$) and does so with high certainty (low entropy). This metric effectively captures the notion that an expert is activated confidently only on specific data domains.


\subsubsection{Driver Expert Identification}
A causal graph \cite{DBLP:journals/corr/abs-1301-2300} represents dependencies among hidden variables in neural networks, facilitating a structured analysis of how internal components affect model outputs. This graph comprises multiple paths from inputs to outputs (i.e., predictions). Our goal is to determine whether certain activated experts have a more critical influence on the prediction process than others.


As shown by \citet{vig2020investigating}, causal mediation analysis naturally quantifies the influence of intermediate variables in causal graphs \cite{DBLP:books/acm/22/BalkeP22}. To measure each activated expert’s contribution to the model’s prediction.
Following the prior work \cite{DBLP:conf/nips/MengBAB22}, we estimate the causal effect of each expert by comparing model outputs from clean and intervened runs.


Given a model input $X$, we investigate the experts in layer $L$, where the expert pool $E = \{e_1, e_2, \cdots, e_M\}$ represents the candidates capable of processing information from $X$. To evaluate the impact of each expert on the model's output, we conduct two types of runs to estimate this effect: \textbf{Clean Run:} The model is executed using the original (unmodified) routing scores. We denote the resulting prediction distribution as $P(X)$ (The process can refer to Equation (1)).
\textbf{Corrupted Run:} To investigate the contribution of individual experts to the final prediction, we apply a unit perturbation (magnitude 1) to the routing scores at the current layer. The model then proceeds with this modified configuration for causal analysis. The resulting prediction distribution is denoted as $Q(X)$.
In the corrupted run, we introduce perturbations to the gating logits before the Softmax operation. Specifically, we modify the routing scores of expert $e_{i}$ as:
\begin{align}
    z = W_g X, \quad
    \tilde{z}= z + \epsilon, \quad
    G_{i}(X)= \mathrm{softmax}(\tilde{z})
\end{align}
where $\epsilon$ is a unit-norm perturbation vector applied to the gating logits. The updated gates are then used to aggregate the expert outputs:
\begin{equation}
        Y = \sum_i G_i(X)\, E_i(X)
\end{equation}
To quantify the sensitivity of model predictions to perturbations of each expert, we measure the change between the clean-run prediction distribution $P(X)$ and the corrupted-run distribution $Q(X)$ using the Kullback--Leibler divergence:
\begin{equation}
    \mathrm{CE}(e_i) = D_{\mathrm{KL}}\!\bigl(P(X)\,\Vert\, Q_i(X)\bigr).
\end{equation}
Here, $\mathrm{CE}$ denotes the causal contribution of expert $e_i$, computed under a unit perturbation and therefore interpretable as a normalized causal effect. In practical sparse MoE architectures, routing distributions are typically highly peaked, with the top-1 or top-2 experts receiving the majority of the routing mass~\cite{du2022glam,lepikhin2020gshard}. Consequently, perturbations to these primary selected experts dominate the weighted output, making KL divergence a reliable empirical proxy for output-level sensitivity even though it is not a complete theoretical causal measure.

\begin{figure*}[!t]
\centerline{\includegraphics[width=\linewidth]{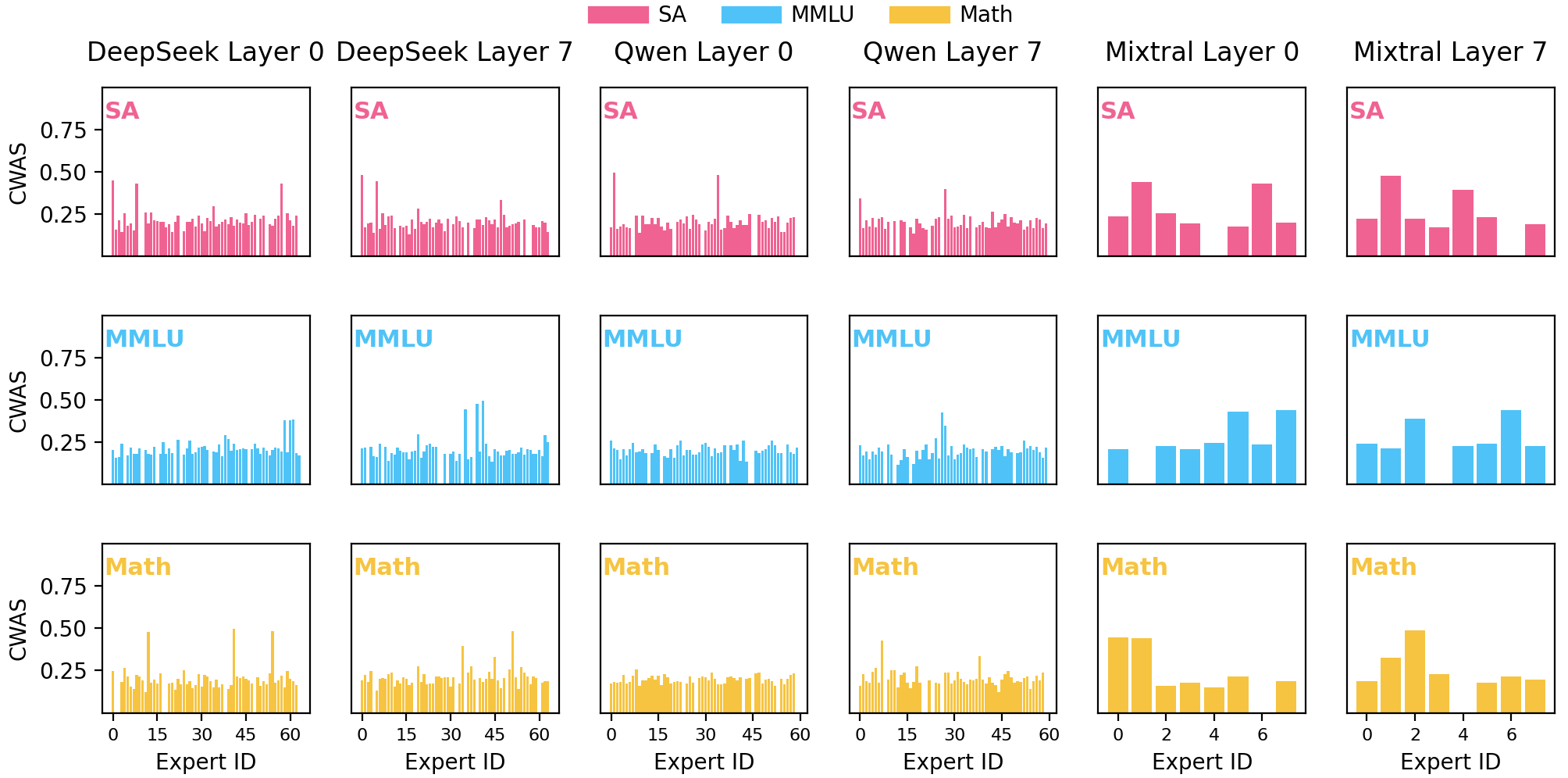}}
\caption{Expert-level Certainty-Weighted Activation Score (CWAS) for DeepSeek-MoE (left), Qwen-MoE (middle), and Mixtral-8×7B (right).}
\label{fig:expert_entropy}
\end{figure*}
\section{Experiments and Result Analysis}
\subsection{Experimental Setting}
This study focuses on three MoE-based LLMs: 1) Mistral \cite{jiang2024mixtral}, 2) DeepSeek-MoE \cite{guo2025deepseek}, and 3) Qwen-MoE \cite{bai2023qwen} and three domains: 1) Sentiment Analysis (SA), 2) Massive Multitask Language Understanding (MMLU) and 3) Math Reasoning. Detailed descriptions are provided below.
\subsubsection{MoE-based LLMs:}
Mixtral\footnote{{https://huggingface.co/mistralai/Mixtral-8x7B}}, DeepSeek-MoE\footnote{{https://huggingface.co/deepseek-ai/deepseek-moe-16b-base}}, and Qwen-MoE\footnote{{https://huggingface.co/Qwen/Qwen1.5-MoE-A2.7B}} adopt sparse Mixture-of-Experts (MoE) architectures, where a subset of expert feedforward networks is dynamically activated per token via a top-$K$ routing mechanism. Specifically, each MoE layer in Mixtral consists of 8 experts with $K=2$ selected per token; DeepSeek-MoE employs 64 experts per layer with $K=6$ activated; and Qwen-MoE uses 60 experts per layer, with $K=4$ activated per token.

\subsubsection{Domains:}
\textbf{Sentiment Analysis (SA):} The task focuses on identifying emotional states \cite{hu2022unimse,hu2024unimeec}. EmotionLines \cite{chen2019emotionlines} is a dialogue emotion classification benchmark built from TV show scripts, with each utterance annotated as angry, disgusted, fearful, happy, sad, surprised, or neutral.
\textbf{Massive Multitask Language Understanding (MMLU):} MMLU \cite{hendrycksmeasuring} is a general-domain benchmark designed to evaluate the reasoning and knowledge capabilities of language models across 57 subjects spanning a wide range of domains.
\textbf{Mathematical Reasoning:} GSM8K \cite{cobbe2021gsm8k} is a widely used benchmark of grade-school math word problems, containing 7,473 train and 1,319 test samples that require multi-step reasoning to yield a single numeric answer. 

\subsection{Q1: Activated Experts Analysis}
To better understand the behavior of activated experts, we examine their token-routing patterns and their influence on model performance from two complementary perspectives: (1) activated expert behavior, and (2) model performance analysis. Detail implementation and results are provided in Appendix.

\subsubsection{Activated Expert Behavior Analysis}
\begin{figure}[!ht]
\centering
\subfigure[\label{fig:visualization:a}]{
  \includegraphics[width=0.41\linewidth]{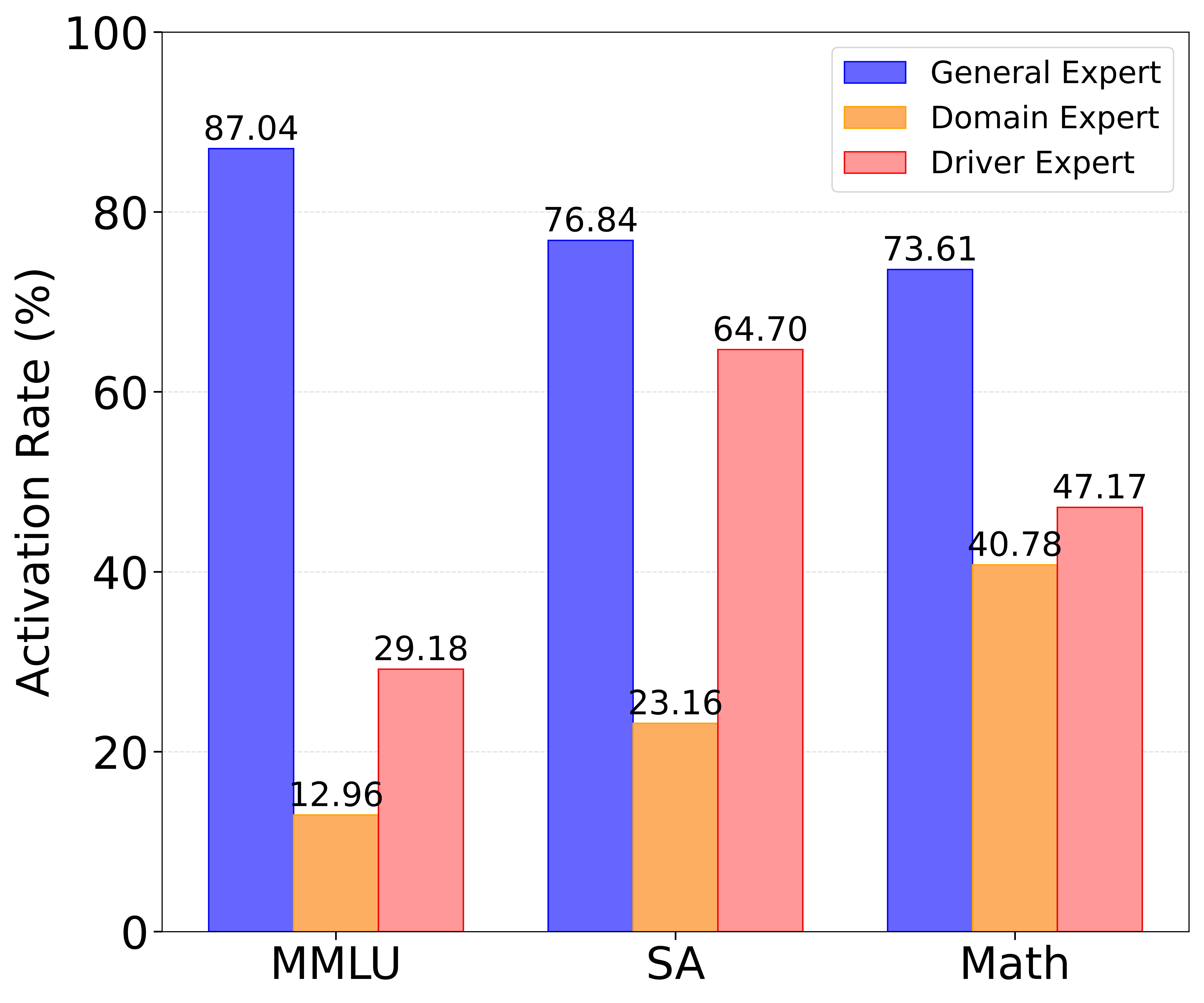}}
\hspace{0.03\linewidth}
\subfigure[\label{fig:visualization:b}]{
  \includegraphics[width=0.41\linewidth]{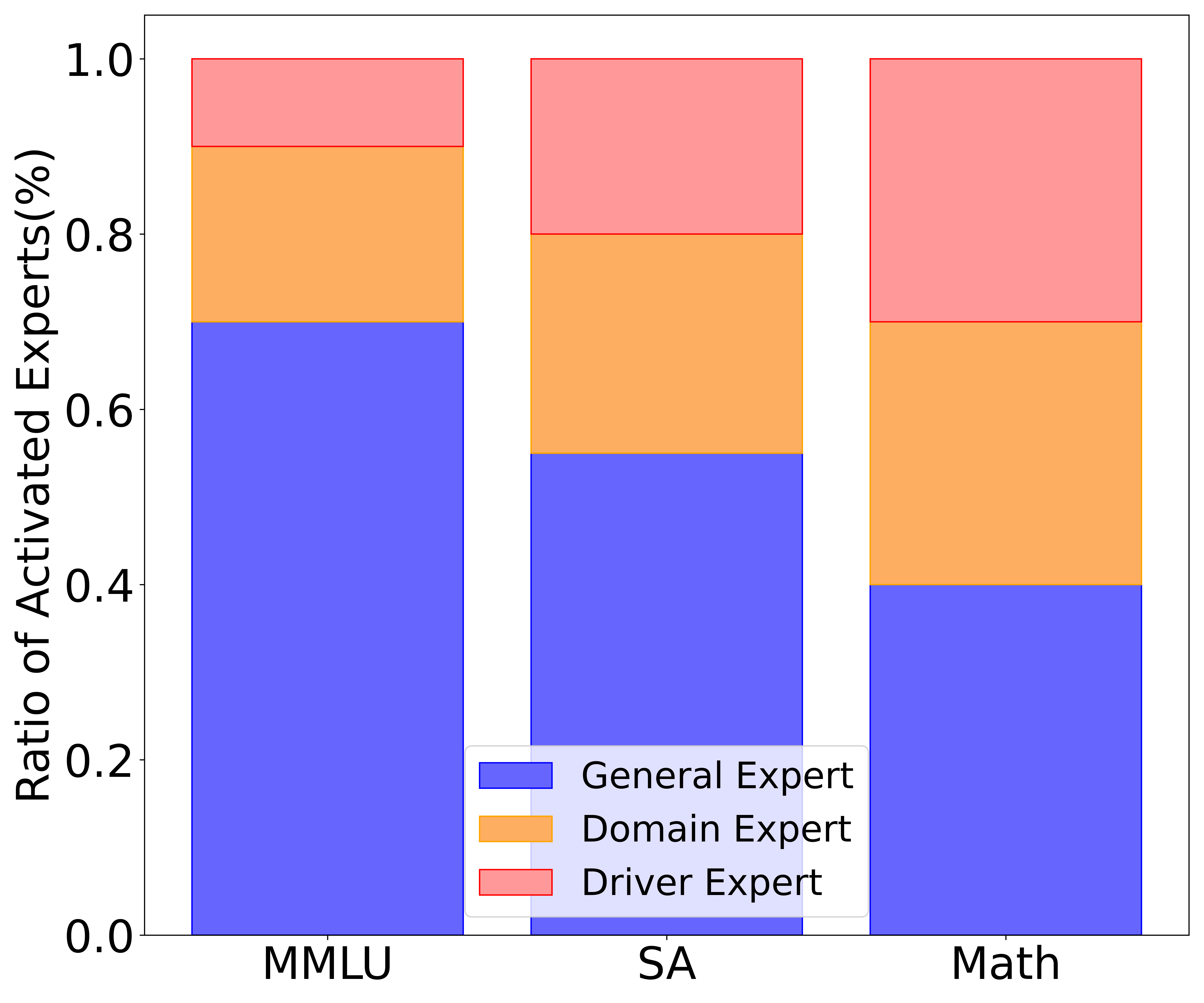}}
\caption{Comparison of expert activation behavior across tasks: (a) activation rates of each expert type, and (b) the distribution of activated experts within each task.}
\label{fig:activated_distribution}
\end{figure}

\paragraph{Expert Activation Behavior Analysis}

We quantify the number of general, domain and driver experts among the activated experts across three corpora: SA, MMLU and Math, as shown in Figure~\ref{fig:activated_distribution}. 
Since domain and driver experts are selected based on separate criteria, the two sets may share some experts. We observe two key findings:
(1) general experts those capturing broad, domain-agnostic knowledge—dominate the learning process across all three domains, whereas domain and driver experts account for only a small fraction of activations;
(2) in the SA and Math domains, however, both domain experts and driver experts contribute much more substantially to model predictions than they do in the MMLU domain. 

\paragraph{Domain Expert}
Figure \ref{fig:expert_entropy} shows clear differences in expert activation patterns across models and domains. DeepSeek and Qwen exhibit mostly flat Certainty-Weighted Activation Score (CWAS) curves with only a few low-CWAS peaks, suggesting moderate specialization and relatively diffuse routing. In contrast, Mixtral displays pronounced low-CWAS experts, indicating stronger specialization despite having fewer experts. Across models, several experts consistently present lower CWAS for specific domains (e.g., Math, MMLU), providing evidence of domain-oriented specialization emerging during pretraining. Overall, the expert-level certainty-weighted activation distribution reveals two key patterns: (1) only a minority of experts displays strong specialization, as indicated by routing frequencies well above the uniform expectation for a domain and (2) most experts exhibit minimal domain-specific behavior, which align well with findings reported in prior work \cite{chaudhari2025moe}.

\paragraph{Drive Expert}
\begin{figure}[t]
\centering
\small
\vspace{4pt}
\includegraphics[width=0.9\linewidth]{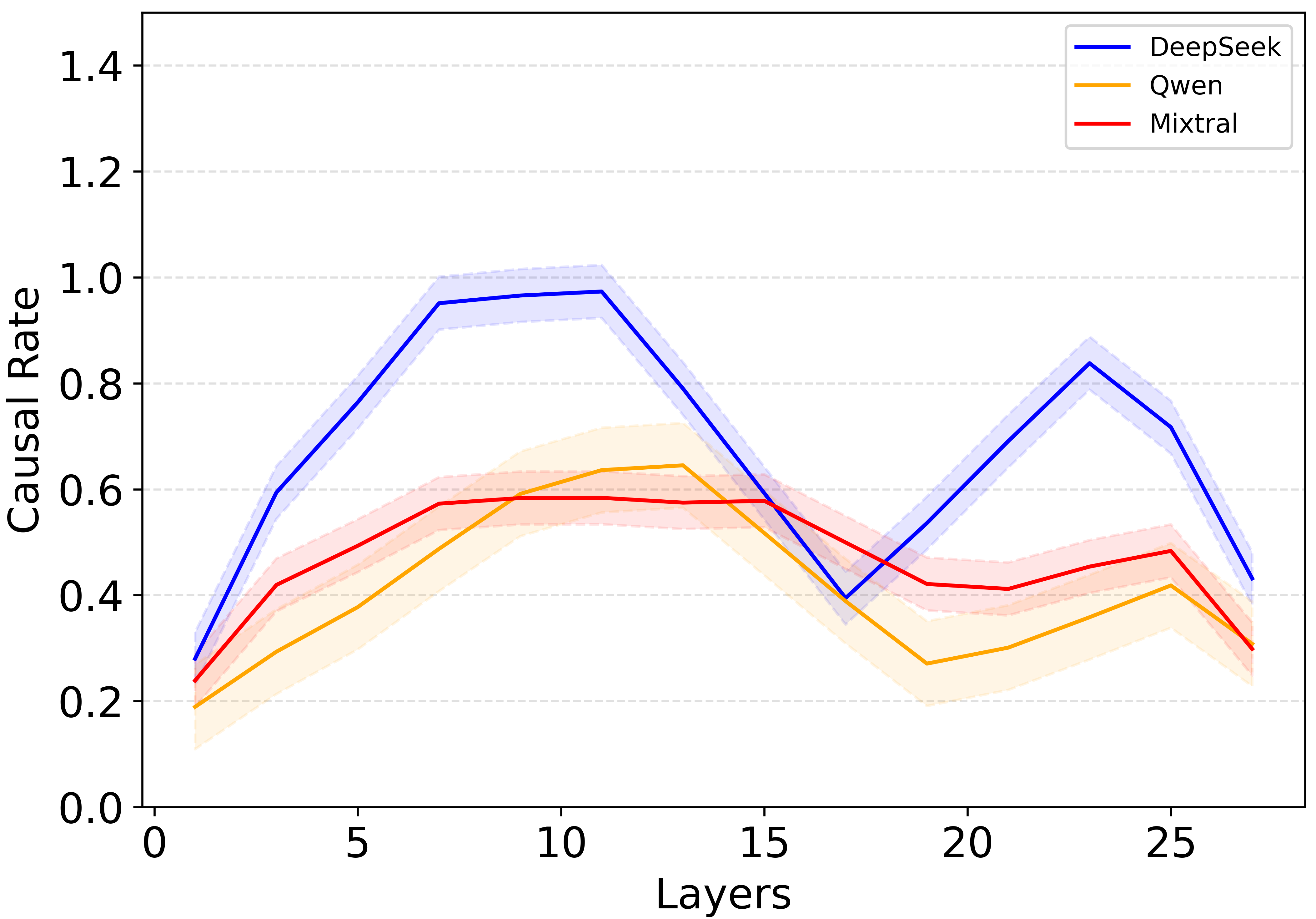}
\captionof{figure}{Layer-wise causal rates.}
\label{fig:causal_rate}
\end{figure}
We further introduce a \textit{causal rate} metric, defined as the fraction of activated tokens that are routed to driver experts across layers in MoE-based LLMs. Higher causal rate means driver experts dominate the model’s decision-making, while lower causal rate reflects more dispersed expert contributions with weaker causal influence. Figure \ref{fig:causal_rate} presents the layer-wise causal rates for three MoE-based LLMs. We observe that the driver experts exist and are positioned within the middle layers.
All three models display a similar trend: causal influence increases in early layers, peaks around the middle of the network, and then drops before rising again in deeper layers. This pattern suggests that mid-layer experts contribute most strongly to model predictions, while early and late layers play more specialized but comparatively weaker roles.
Across all layers, DeepSeek consistently exhibits the highest causal rate, indicating causality influence on the model’s output. Qwen shows moderate causal sensitivity, while Mixtral maintains the lowest causal rates throughout most layers. More analysis can be referred to Appendix \ref{sec:activation}.

\subsubsection{Model Performance Analysis}
To further demonstrate their roles in model performance, we analyze the impact of activated experts on SA and Math domains by increasing ($\uparrow$) or decreasing ($\downarrow$) their routing weights, focusing specifically on activated domain and driver experts. The results are presented in Table \ref{tab:merged_moe_performance} and Table \ref{tab:moe_performance_sa}. 

\paragraph{Increasing expert weights improves performance:}
During the forward pass, the MoE router emits a logit for each expert; a softmax followed by top-$K$ selection activates the experts. Before the softmax, we up-weight targeted experts (e.g., domain experts) using a scaling factor (e.g., 1.2). Increasing the weights of both the domain expert and the driver expert yields stable performance gains, confirming their importance in adapting to the Math domain.
Increasing the contribution of either domain or driver experts consistently improves performance on SA and Math across all three MoE-based LLMs. In SA, emphasizing domain experts raises DeepSeek’s accuracy from 0.6080 to 0.6243, while emphasizing driver experts yields even larger gains. In Math, boosting either domain or driver experts likewise improves performance for all three models. These findings underscore the complementary roles of domain and driver experts: domain experts contribute contextual grounding and semantic precision, while driver experts provide critical reasoning signals that are especially beneficial in tasks requiring causal inference.

\paragraph{Decreasing expert weights decrease performance:}
Analogous to up-weighting, targeted experts (e.g., domain experts) can be down-weighted by multiplying their logits by a factor < 1 (e.g., 0.8) before the softmax. As expected, reducing expert weights generally degrades performance, with the largest drops observed when the driver expert is down-weighted in the SA and Math domains. For instance, reducing the domain expert's contribution leads to substantial drops: Mixtral’s accuracy declines from 0.5137 to 0.4307, DeepSeek’s from 0.6080 to 0.5559, and Qwen’s weighted F1 from 0.6134 to 0.5416. The impact is even more pronounced when driver experts are down-weighted: Mixtral’s accuracy drops to 0.4411, DeepSeek’s to 0.5146, and Qwen’s weighted F1 to 0.5030. Conversely, decreasing these weights leads to a clear drop in accuracy, indicating that weakening domain-relevant expertise harms model performance.
These results underscore the critical role of both domain-specific and driver experts in preserving model performance and suggest that their influence should be maintained or enhanced during expert routing. 
\begin{table*}[t]
\centering
\resizebox{0.89\textwidth}{!}{
\begin{tabular}{l|ccc|ccc}
\toprule
& \multicolumn{3}{c|}{\bf Math Domain} & \multicolumn{3}{c}{\bf MMLU Domain} \\
\cmidrule(lr){2-4} \cmidrule(lr){5-7}
{\bf Models} 
& {\bf Mixtral} & {\bf DeepSeek} & {\bf Qwen}
& {\bf Mixtral} & {\bf DeepSeek} & {\bf Qwen} \\
\toprule

Original Expert ($-$) 
& 0.8013 & 0.8462 & 0.8117 
& 0.6323 & 0.4761 & 0.6011 \\

Domain Expert ($\uparrow$)
& 0.8209(\textcolor{green}{$\uparrow$}) 
& 0.8676(\textcolor{green}{$\uparrow$}) 
& 0.8162(\textcolor{green}{$\uparrow$})
& 0.6416(\textcolor{green}{$\uparrow$}) 
& 0.4870(\textcolor{green}{$\uparrow$}) 
& 0.6112(\textcolor{green}{$\uparrow$}) \\

Domain Expert ($\downarrow$)
& 0.7814(\textcolor{red}{$\downarrow$}) 
& 0.8128(\textcolor{red}{$\downarrow$}) 
& 0.8040(\textcolor{red}{$\downarrow$})
& 0.6298(\textcolor{red}{$\downarrow$}) 
& 0.4549(\textcolor{red}{$\downarrow$}) 
& 0.5896(\textcolor{red}{$\downarrow$}) \\

Driver Expert ($\uparrow$)
& 0.8209(\textcolor{green}{$\uparrow$}) 
& 0.8676(\textcolor{green}{$\uparrow$}) 
& 0.8162(\textcolor{green}{$\uparrow$})
& 0.6368(\textcolor{green}{$\uparrow$})
& 0.4789(\textcolor{green}{$\uparrow$}) 
& 0.6106(\textcolor{green}{$\uparrow$}) \\

Driver Expert ($\downarrow$)
& 0.7522(\textcolor{red}{$\downarrow$}) 
& 0.7581(\textcolor{red}{$\downarrow$}) 
& 0.7503(\textcolor{red}{$\downarrow$})
& 0.6247(\textcolor{red}{$\downarrow$})
& 0.4701(\textcolor{red}{$\downarrow$}) 
& 0.5914(\textcolor{red}{$\downarrow$}) \\

Finetuning Router
& 0.8176(\textcolor{green}{$\uparrow$}) 
& 0.8488(\textcolor{green}{$\uparrow$}) 
& 0.8214(\textcolor{green}{$\uparrow$})
& 0.6383(\textcolor{green}{$\uparrow$})
& 0.4885(\textcolor{green}{$\uparrow$}) 
& 0.6166(\textcolor{green}{$\uparrow$}) \\

\bottomrule
\end{tabular}}
\caption{The effect of increasing ($\uparrow$) or decreasing ($\downarrow$) expert routing weights across Math and MMLU domains.}
\label{tab:merged_moe_performance}
\end{table*}

\paragraph{Finetuning MoE router only:}
We refine each expert’s router weight by applying a LoRA adaptation to the gating linear projection. Specifically, for MoE-based LLMs (e.g., Mixtral, DeepSeek, and Qwen), we attach LoRA \cite{hu2022lora} to the gating projection in the last three Transformer layers according to the prior work \cite{zoph2022st}. The adaptation is realized via hook-based interception of the projection’s forward pass: a LoRA term is composed with the native logits to adjust router weights without altering the underlying module. The backbone and expert parameters remain frozen, and routing follows the standard softmax top-k with fixed temperature and capacity. We observe that fine-tuning expert weights improves DeepSeek and Qwen compared with the original assignment, whereas Mixtral sees diminished. The Finetuning Router achieves moderate improvements, though its effectiveness varies across models.
In all cases, the effect is smaller than that of upweighting specific experts (e.g., domain or driver experts).

These findings highlight the potential of domain-aware expert activation, where reasoning-centric tasks (e.g., math domain) benefit from driver experts and domain-focused knowledge (e.g., MMLU domain) from domain experts. 

\subsection{Q2: Expert-Activated Token Analysis}
To gain deeper insight into token–expert interactions, we analyze how specific tokens are routed to experts during inference from two perspectives:(1) the sentence-position distribution of tokens that activate experts, and (2) representative tokens associated with domain and driver experts. Additional implementation details are provided in Appendix~\ref{appendix:implementation}.

\begin{table*}[!ht]
\centering
\resizebox{0.89\textwidth}{!}{
\begin{tabular}{lcccccc}
\toprule
                           & \multicolumn{2}{c}{\bf Mixtral} & \multicolumn{2}{c}{\bf DeepSeek} & \multicolumn{2}{c}{\bf Qwen} \\
{\bf Models}                  & {\bf ACC}       & {\bf WF1}       & {\bf ACC}        & {\bf WF1} & {\bf ACC}        & {\bf WF1}        \\
\toprule
Original Expert($-$) &0.5137 & 0.5206 &0.6080 &0.6110 &0.6134&0.6159\\
Domain Expert($\uparrow$) &0.5460(\textcolor{green}{$\uparrow$}) &0.5492(\textcolor{green}{$\uparrow$}) &0.6243(\textcolor{green}{$\uparrow$}) &0.6334(\textcolor{green}{$\uparrow$})&0.6250(\textcolor{green}{$\uparrow$}) &0.6288(\textcolor{green}{$\uparrow$})\\
Domain Expert($\downarrow$) &0.4411(\textcolor{red}{$\downarrow$}) &0.4447(\textcolor{red}{$\downarrow$}) &0.5528(\textcolor{red}{$\downarrow$}) &0.5559(\textcolor{red}{$\downarrow$})&0.5416(\textcolor{red}{$\downarrow$}) &0.5572((\textcolor{red}{$\downarrow$})\\
Driver Expert({$\uparrow$}) &0.5685(\textcolor{green}{$\uparrow$}) &0.5842(\textcolor{green}{$\uparrow$})&0.6442(\textcolor{green}{$\uparrow$}) &0.6481(\textcolor{green}{$\uparrow$})&0.6432(\textcolor{green}{$\uparrow$}) &0.6512(\textcolor{green}{$\uparrow$})\\
Driver Expert({$\downarrow$}) &0.4307(\textcolor{red}{$\downarrow$}) &0.4336(\textcolor{red}{\textcolor{red}{$\downarrow$}}) &0.5146(\textcolor{red}{$\downarrow$}) &0.5202(\textcolor{red}{$\downarrow$})&0.5030(\textcolor{red}{$\downarrow$}) &0.5122(\textcolor{red}{$\downarrow$})\\
Finetuning Router&0.5007(\textcolor{red}{$\downarrow$})&0.5125(\textcolor{red}{$\downarrow$})&0.6102(\textcolor{green}{$\uparrow$})&0.6129(\textcolor{green}{$\uparrow$})&0.6145(\textcolor{green}{\textcolor{green}{$\uparrow$}})&0.6170(\textcolor{green}{$\uparrow$})\\
\bottomrule
\end{tabular}}
\caption{The effect of ajusting expert routing weights compared to original expert (denoted by $-$) on SA domain.}
\label{tab:moe_performance_sa}
\end{table*}
\begin{figure*}[t]
\centering

\subfigure[SA\label{fig:visualization:a}]{
\includegraphics[width=0.30\linewidth]{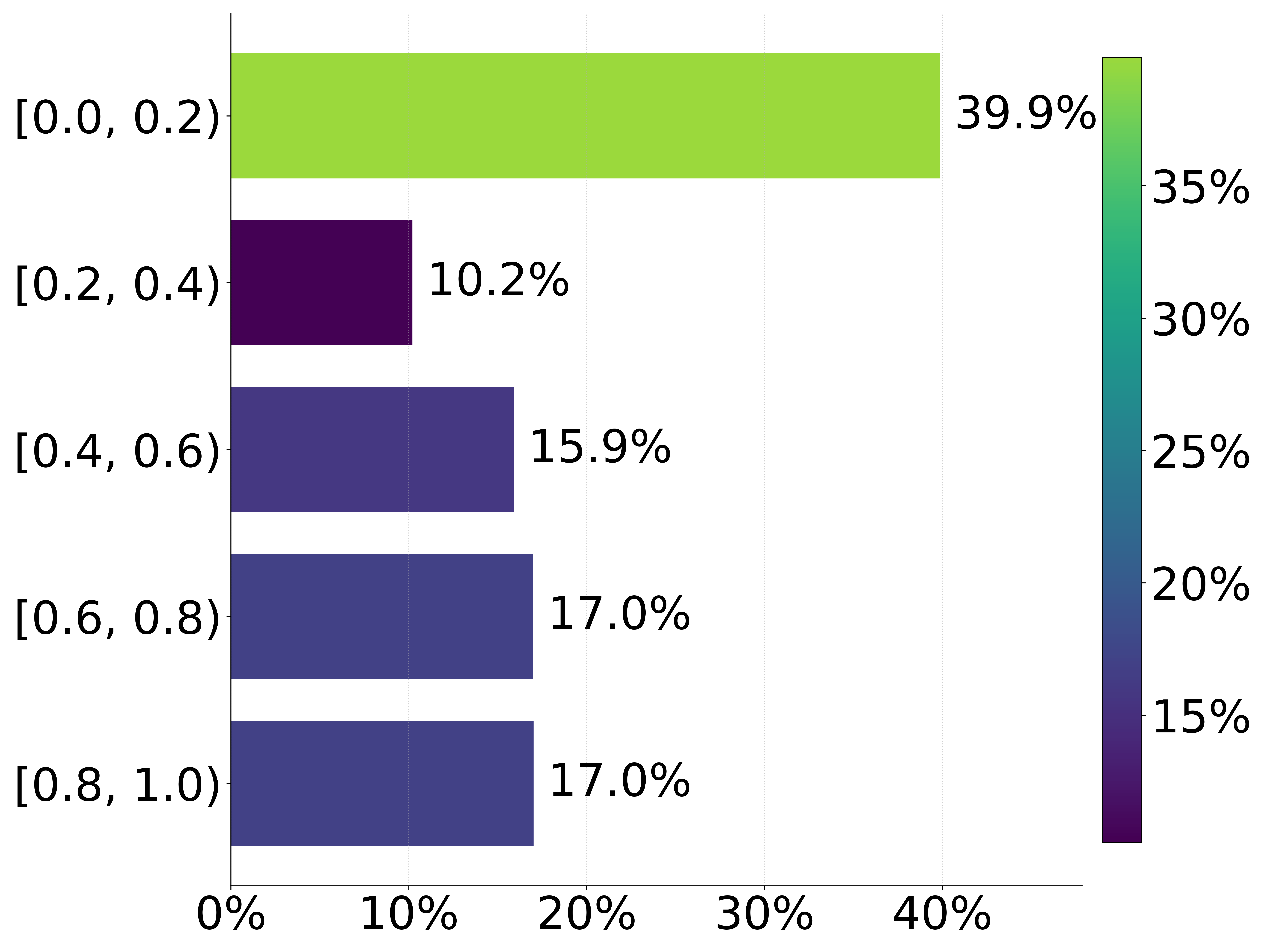}
}
\hfill
\subfigure[MMLU\label{fig:visualization:b}]{
\includegraphics[width=0.30\linewidth]{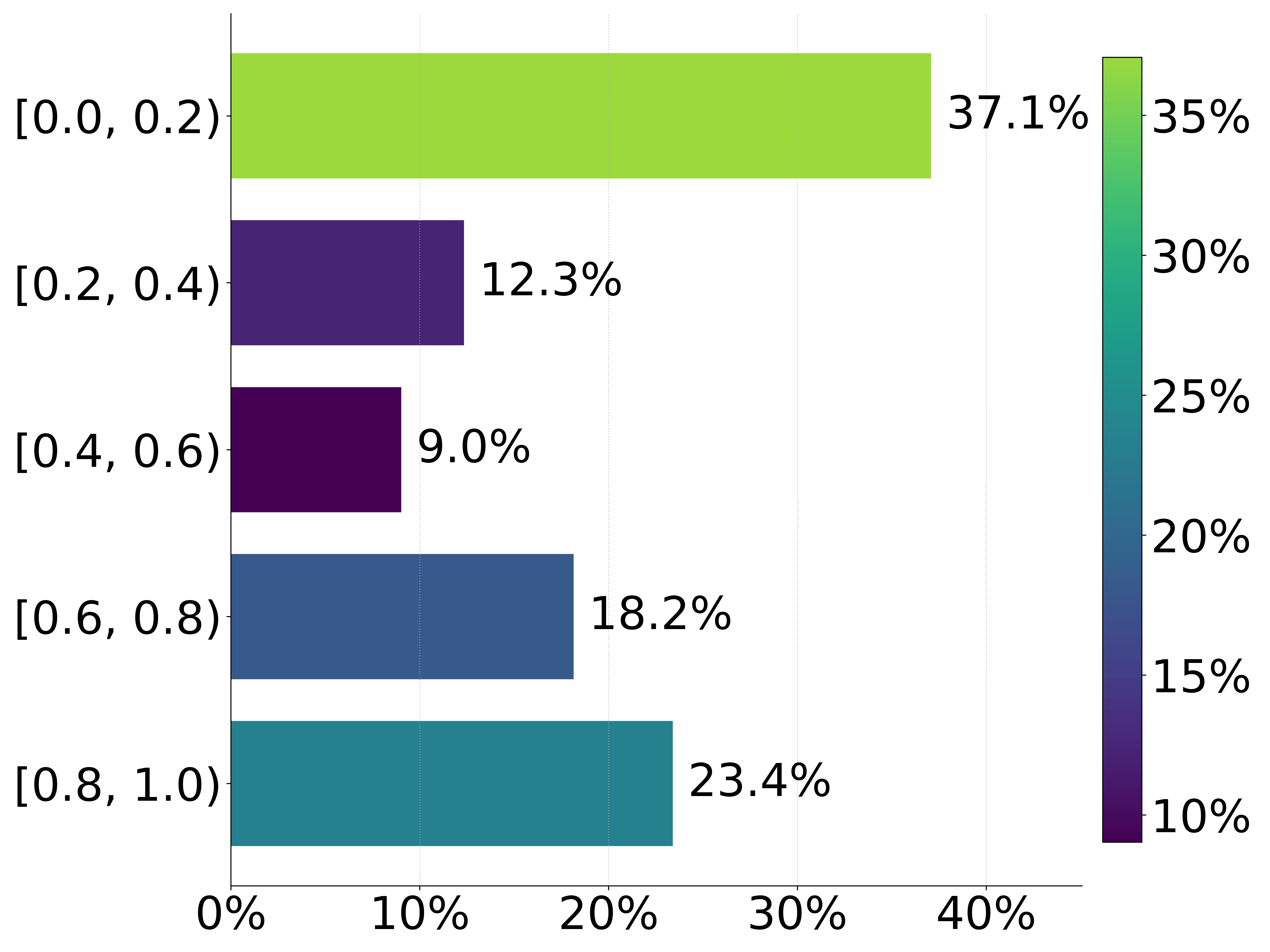}
}
\hfill
\subfigure[Math\label{fig:visualization:c}]{
\includegraphics[width=0.30\linewidth]{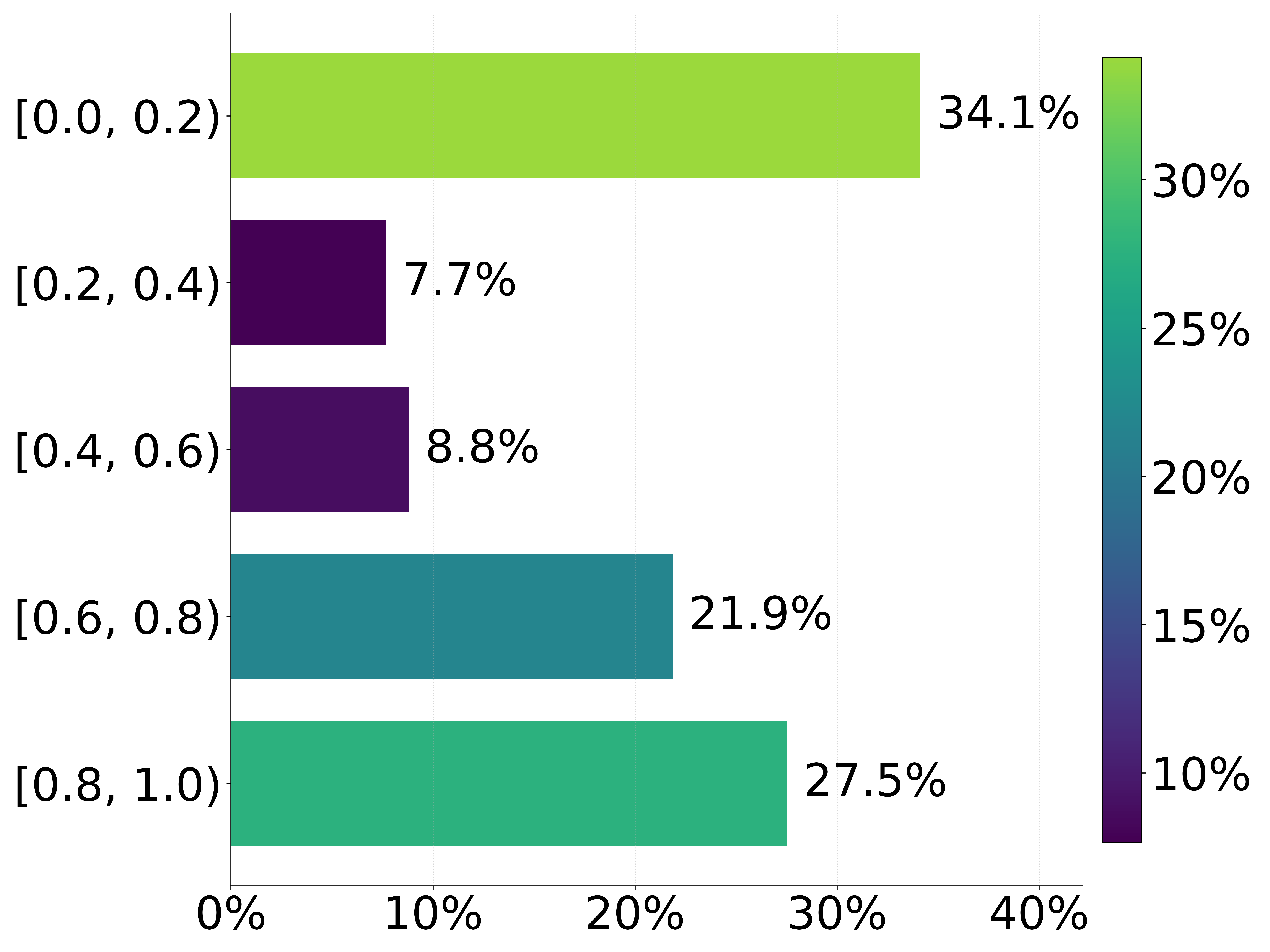}
}

\caption{
Proportional causal influence of tokens by sentence position under the DeepSeek model. 
The horizontal bar charts illustrate token contributions across sentence positions for 
three domains (SA, MMLU, and Math).
}
\label{fig:heatmap_bar}
\end{figure*}

\subsubsection{Distribution of Expert-Activated Tokens}
To analyze where expert-activating tokens occur within a sentence, we divide each input sequence into five equal-length positional bins: [0.0,0.2),[0.2,0.4),[0.4,0.6),[0.6,0.8),[0.8,1.0). For each token, we collect its top-K routed experts and count whether any driver experts are activated. Tokens that activate at least one driver expert are retained. We then compute the proportion of these tokens within each positional bin and visualize their distribution for the SA and MMLU domains, as shown in Figure~\ref{fig:heatmap_bar}. Both charts indicate that tokens at the start of the sentence carry disproportionately high causal weight in model predictions, which aligns to prior work \cite{barbero2025llms}. This may reflect a left-to-right autoregressive bias (as in GPT-style models), or the prominence of introductory context. The central part of the sentence contributes the least, possibly because these tokens carry less contextual or structural salience in many sentences.
Additionally, the latter part of the sentence also exhibits strong causal effects on model predictions \cite{DBLP:conf/nips/MengBAB22}. This phenomenon can be attributed to the fact that the latter portion of a sentence frequently encompasses critical semantic information that plays a decisive role in guiding the model’s predictions. Also, in autoregressive frameworks, tokens appearing later in the sequence possess greater causal influence due to their temporal proximity to the prediction point, thereby exerting stronger effects on the model’s output.


\subsubsection{Tokens Associated with Experts}
\begin{figure}[t]
\centering
\small
\vspace{1pt}
\resizebox{0.9\linewidth}{!}{\begin{tabular}{llc}
\toprule
{\bf Domain} & {\bf Expert} & {\bf Token} \\
\toprule
\multirow{2}{*}{SA} & domain & dumped,fossil,hot,clean,food \\
                    & driver & dumped, crazy,panic,jealous,theory \\
\multirow{2}{*}{MMLU} & domain & revolution,hypertension,treaty \\
                    & driver & what, how, because, therefore \\
\multirow{2}{*}{Math} & domain & average, multiple, percent, sum \\
                    & driver & sum, minimum, average, maximum \\
\midrule
\end{tabular}}
\captionof{table}{Tokens associated with domain and driver experts.}
\label{tab:domain_tokens}   
\end{figure}
For each token $x_{t}$, we extract the top-k expert assignments from the router and aggregate routing statistics across all occurrences of the token within each evaluation dataset. Domain and driver experts are identified using the criteria described earlier (entropy-based and causal-effect-based). We then compute token–expert association scores and report representative examples in Table ~\ref{tab:domain_tokens}.

Each task is associated with both domain and driver experts. For example, the EmotionLine corpus of SA is constructed from TV show Friends. The domain focuses on daily activities along with the traits of each character. Specially, tokens such as ``dumped'', ``fossil'', ``hot'', ``clean'', and ``food'' are topic-specific and often carry implicit sentiment—dumped may relate to relationships, while hot can refer to physical appearance. These tokens typically activate the domain expert, which specializes in leveraging semantic and contextual understanding. In contrast, tokens like ``dumped'', ``crazy'', ``panic'', and ``jealous'' indicate emotional causes or consequences. Such cues engage the driver expert, enabling the model to reason about emotional dynamics and underlying triggers. For MMLU and Math, the tokens associated with domain and driver experts differ markedly from those in SA. This contrast reflects the natures of domain and driver experts: driver experts are triggered by interrogatives (e.g., ``what'', ``how''), whereas domain experts respond to specialized vocabulary (e.g., ``revolution'', ``treaty''). In the Math setting, domain experts attend to technical terms (e.g., ``percent'', ``multiple''), while driver experts are activated by operation-defining tokens (e.g., ``sum'', ``minimum''), i.e., sensitivity cues. Compared to tokens associated with domain experts, which are typically content words, those linked to driver experts are more often function words. This insight provides a valuable foundation for designing more effective and task-adaptive expert routing strategies that leverage linguistic features to improve both model efficiency and transparency.

\section{Related Work}
\subsection{Advances in Mixture-of-Expert}
Mixture-of-Expert(MoE) architectures have been successfully integrated into large-scale models, particularly in natural language processing (NLP) and computer vision (CV) \cite{lin2024moe,bao2022vlmo}. \citet{riquelme2021scaling} applies MoE to Vision Transformers (ViTs) and \citet{yu2023mmoe} proposes multimodal mixtures of experts (MMOE) to train separate expert models for each type of multimodal interaction. More recently, \citet{DBLP:journals/corr/abs-2503-07639} proposes MoE-X, an interpretable mixture of experts large language model, while maintaining competitive performance. \citet{lo2025closer} quantifies expert activation/routing and show routers favor high-norm experts while specialization/diversity grow with depth. OLMoE \cite{muennighoffolmoe} releases fully open sparse-MoE LLMs and also offers thorough ablations on routing, auxiliary losses, and dense-to-MoE upcycling. \citet{fayyaz2025steering} proposes SteerMoE to identify behavior-related experts in MoE models and dynamically activates or deactivates them at inference time to steer model trustworthiness and safety, without requiring retraining.
\citet{bandarkar2025multilingual} analyze multilingual token routing in MoE models using parallel corpora, revealing systematic layer-wise expert selection patterns that inform routing design for multilingual LLMs.

\subsection{Advances in Mechanistic Interpretability}
Mechanistic interpretability seeks to explain how internal circuits transform inputs into predictions, with prior work primarily focusing on the functions of attention heads across different settings \cite{DBLP:conf/nips/MengBAB22,DBLP:conf/iclr/ToddLSMWB24}. For instance, \citet{yu2024neuron} investigates how individual neurons affect output distributions. Furthermore, works \cite{DBLP:conf/acl/TangLH0WZWW24} and \cite{DBLP:conf/emnlp/HuoYHYH24} identify language-specific and domain-specific neurons respectively for better utilize language and domain information. Recently, lots of work have focused on the understanding of expert specialization, finding that different experts are responsible for different tokens, domains \cite{xue2024openmoe,yang2025qwen3}, tasks \cite{fayyaz2025steering}. These findings facilitate the targeted optimization of the specialized experts \cite{chaudhari2025moe}.
However, MoE interpretability gap: (i) which experts activate (and how consistently by type); (ii) which tokens trigger which experts. We instead target domain and driver experts to uncover activation mechanism in MoE LMs.

\section*{Conclusion}
Our work aim to address two central questions in MoE LMs: (a) Which experts are activated? and (b) How are input tokens associated with specific experts? We propose entropy-based and causal effect methods to identify domain and driver experts, and further analyze the association between tokens and activated experts.
Our analysis reveals two key insights: (1) activated experts fall into two distinct types, domain and driver, and (2) tokens at the beginning of a sentence are more likely to trigger driver experts, and (3) adjusting the weights of domain and driver experts yields improvements in accuracy and F1 across all MoE-based LLMs and domains, respectively. These findings illuminate MoE internals and interpretability, offering a fresh lens for integration into existing architectures.

\section*{Limitations}
We study three models and three domains to probe the mechanisms of activated experts and to verify the presence of domain and driver experts. Identification relies on thresholds and routing logs (e.g., top-k gating, load balancing), so our results may be sensitive to these design choices and hyperparameters. We evaluate only three MoE LLMs; architectural and scale differences (routing strategy, expert capacity, training recipe) may affect the prevalence and behavior of domain/driver experts.

\bibliography{acl_latex}

\appendix

\section{More Related Work}

\subsection{Mixture of Experts}
Mixture-of-Experts (MoE) architectures activate only a small subset of experts (e.g., 2 to 6) for each input token, enabling efficient scaling to models with hundreds of billions of parameters. Each token in the input sequence is independently routed by a lightweight gating network, which outputs a probability distribution over a fixed set of expert sub-networks. The model then activates the top‑$K$ experts with the highest gating scores to process each token. In practice, the MoE module is integrated into the Feed-Forward Network (FFN) layers of the transformer, where each FFN functions as an expert. During inference, the gating function $G(x)$ controls expert activation by computing a softmax over the dot product between the input token representation $H^i$ and the weights $W_g$, resulting in a distribution over experts $G(x) \in \mathbb{R}^N$. The final output $H_F^{i+1}$ is then obtained by weighting the outputs of the activated experts $e_i(x)$ according to the gating scores $G_i(x)$.
\begin{equation}
\begin{split}
    &G(x)=softmax(W_{g}\cdot H^{i})\\
    &H^{i+1}_{F} = G_{i}(x)\cdot e_{i}(x)
\end{split}
\end{equation}
where $e_i$ represents the output (or information) from the $i$-th expert module.

\begin{figure*}[t]
\centerline{\includegraphics[width=\linewidth]{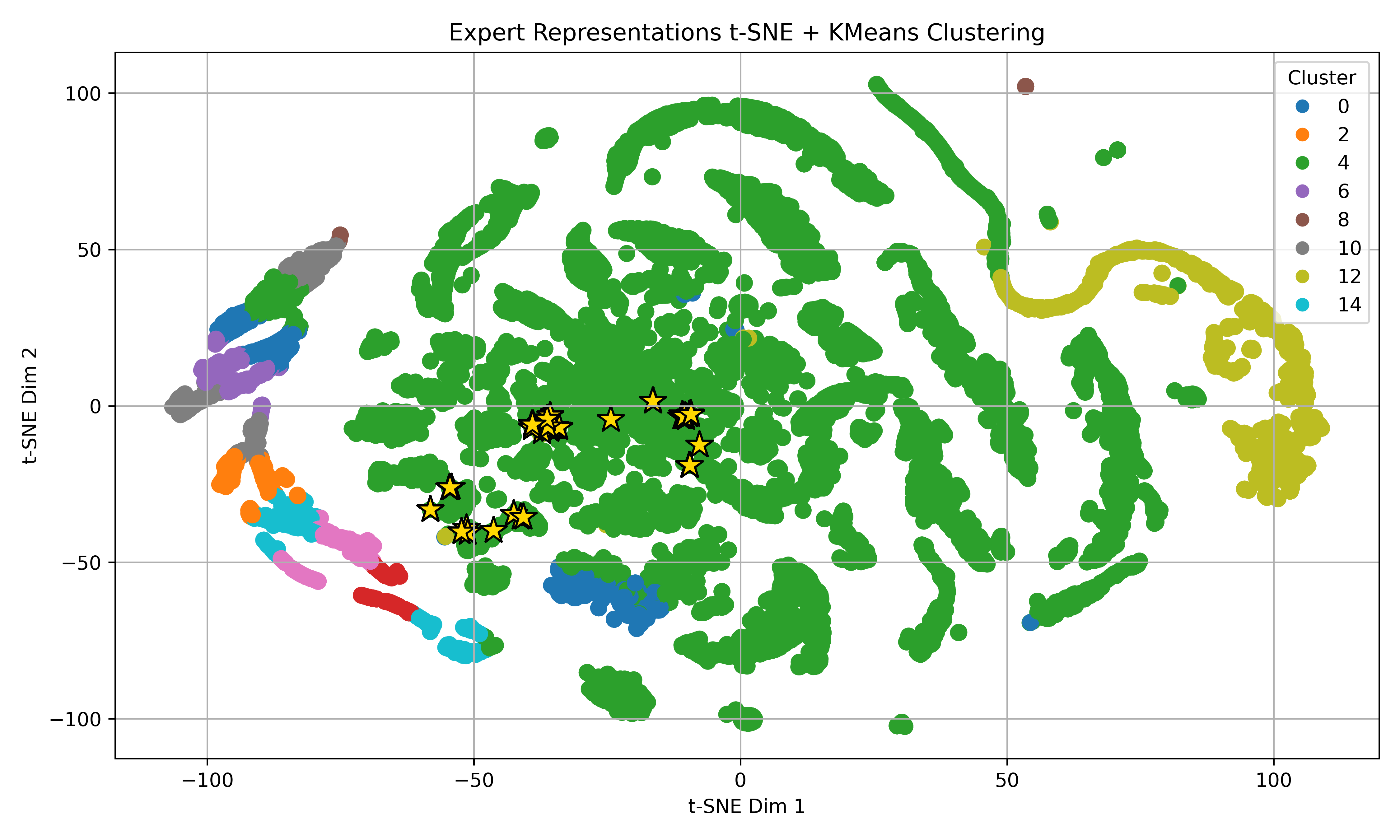}}
\caption{T-SNE visualization of expert representations with KMeans clustering. Driver experts (yellow stars) form a compact and coherent subregion in the embedding space, indicating shared functional characteristics and high causal influence.}
\label{fig:expert_representation}
\end{figure*}

\section{More Experimental Results}
Figure \ref{fig:expert_representation} visualizes the expert representations using t-SNE followed by K-Means clustering. We observe several notable patterns.

The embedding space naturally separates into multiple coherent clusters and suggests that MoE experts learn diverse specialization patterns during pretraining.
First, most experts exhibit highly homogeneous representations, with only a small subset learning diverse and specialized behaviors. This pattern aligns with observations reported in \cite{}, suggesting that MoE models tend to rely on a limited number of experts for specialized computation while the majority remain broadly general.

\begin{figure}
\centering
\includegraphics[width=0.98\linewidth]{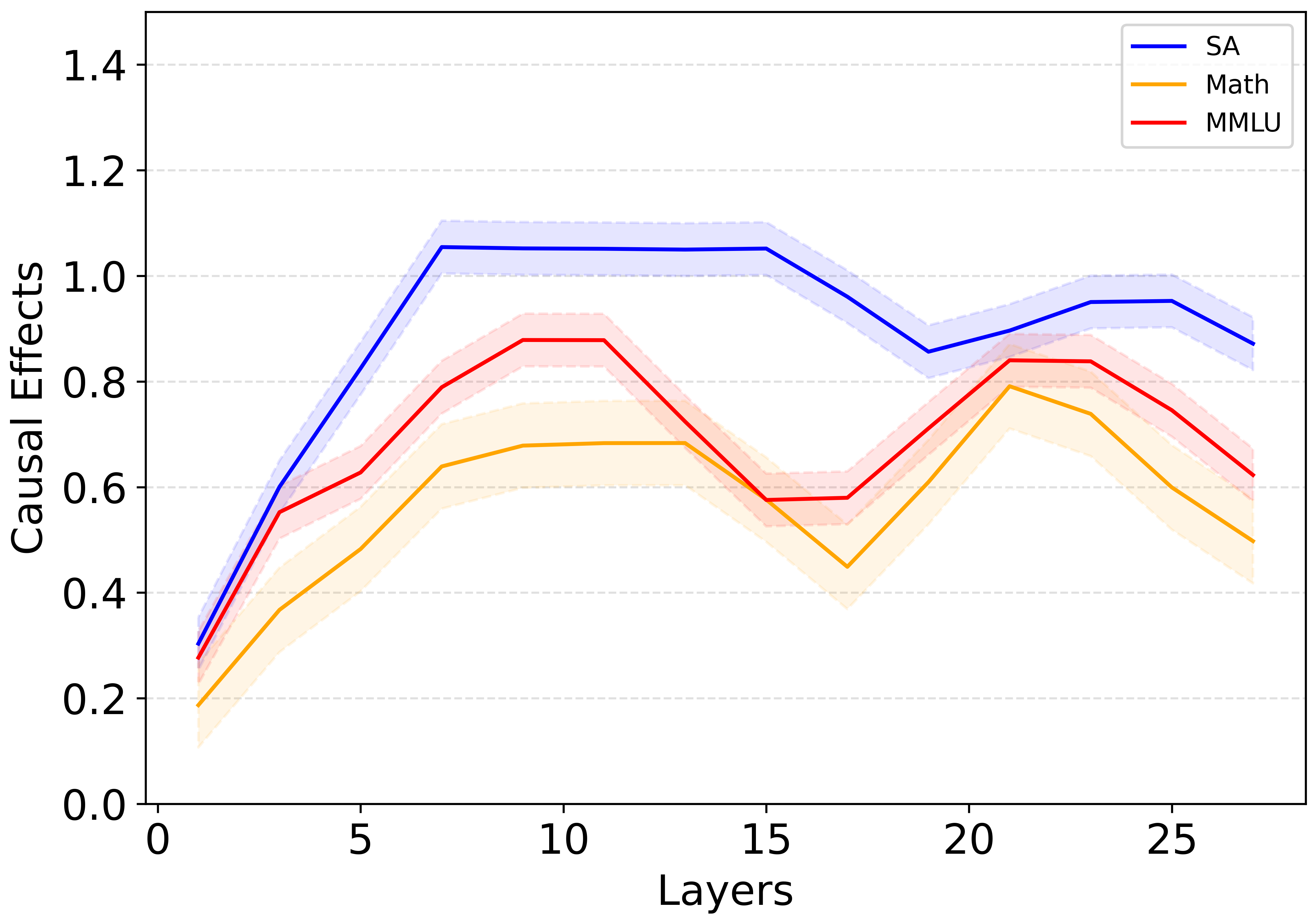}
\captionof{figure}{Layer-wise causal effects.}
\label{fig:causal_effects}
\end{figure}

Second, driver experts (highlighted with yellow stars) do not uniformly scatter across all clusters. Instead, they form a concentrated subregion within the embedding space. This spatial coherence implies that driver experts share similar functional characteristics, consistent with their high causal contributions in earlier analyses. Their tight grouping also suggests that driver experts may capture a common underlying computation—likely involving cross-domain reasoning or high-influence transformations.

Overall, the visualization reveals that (1) expert representations are structurally organized into meaningful clusters, and (2) driver experts occupy a compact and distinct region in the embedding space. These findings provide strong evidence that expert behavior in MoE models is both structured and functionally specialized.
\paragraph{Drive Expert}

Figure \ref{fig:causal_effects} illustrates the layer-wise causal effects of experts across three domains: SA, Math, and MMLU
Overall, all three curves follow a similar trend—causal influence rises sharply in the early layers, peaks around the middle layers (approximately layers 8–12), and then gradually stabilizes. This pattern indicates that mid-layer experts play a disproportionately important role in shaping the model’s predictions across domains.
Among the three tasks, SA consistently exhibits the highest causal effects, suggesting that the model relies more heavily on its driver experts when processing emotionally rich or context-dependent inputs. Math shows moderate causal contributions, while MMLU remains the lowest across most layers, indicating that reasoning-heavy or knowledge-intensive tasks distribute influence more evenly across experts.
Notably, the sharper rise and stronger peak in the SA curve imply that its driver experts exert greater and more concentrated causal impact, whereas Math and MMLU demonstrate smoother and more diffused expert contributions. These findings collectively suggest that different domains trigger distinct expert activation dynamics, and mid-layer experts are particularly critical regardless of the task.

\section{Implementation Details}
\label{appendix:implementation}
\subsection{Experimental Settings}
All experiments are inference-only on NVIDIA RTX A100 and RTX A40 GPUs. Domain- and driver-expert identification is performed with frozen parameters. For expert-weight interventions, we up- or downweight the identified experts while keeping all other parameters fixed. Both expert-entropy and causal-effect computations are performed at inference, incurring no additional training cost. In experiments assessing expert impact, we modify only the router weights while keeping all model parameters frozen.

\section{Activated Expert Analysis}
\label{sec:activation}

\begin{figure*}[t]
\centerline{\includegraphics[width=0.85\textwidth]{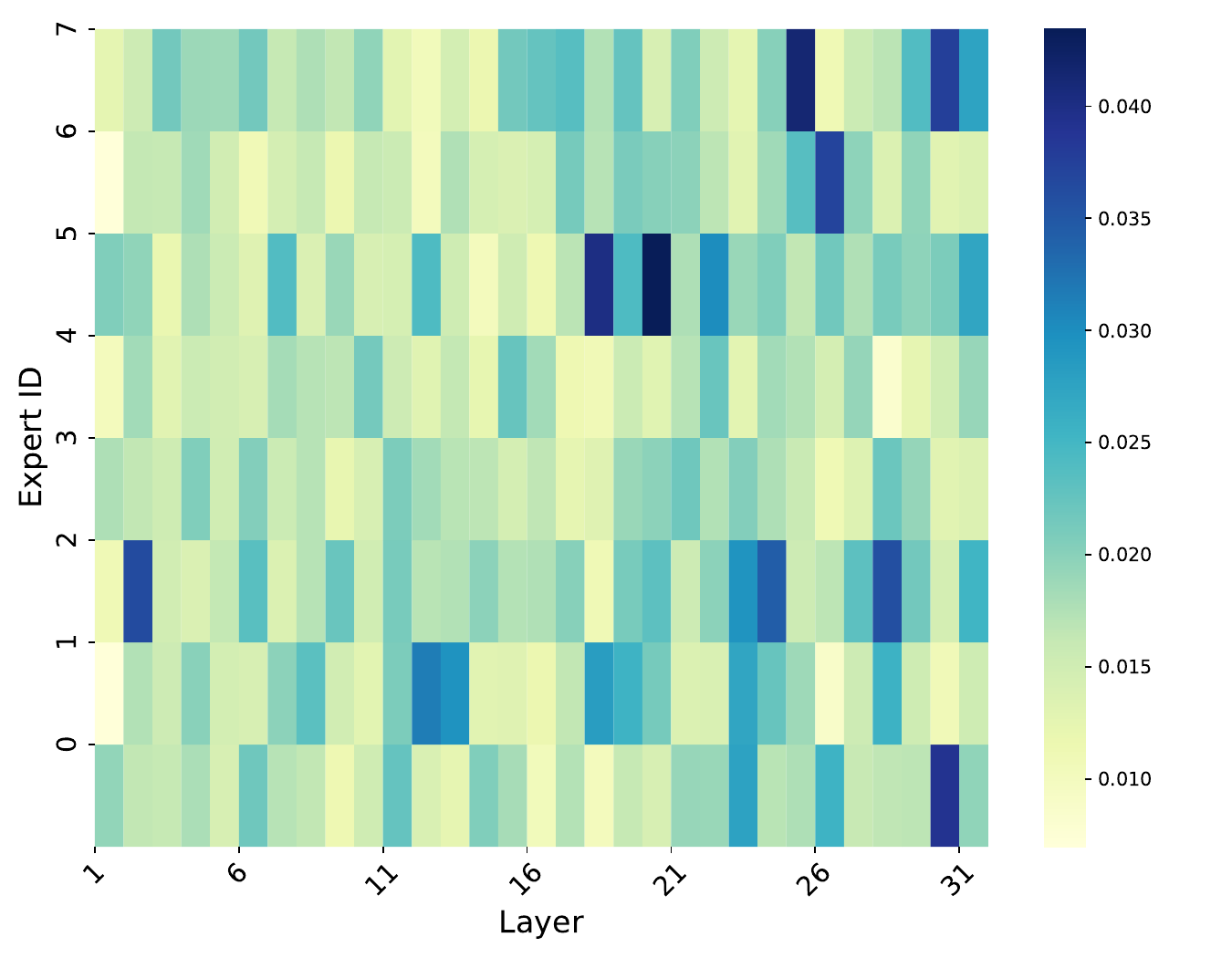}}
\caption{Heatmap of expert-level causal influence across layers for DeepSeek model. Darker colors indicate stronger causal effects on the model’s output.}
\label{fig:causal_impact_deepseek}
\end{figure*}

\begin{figure*}[t]
\centerline{\includegraphics[width=0.9\textwidth]{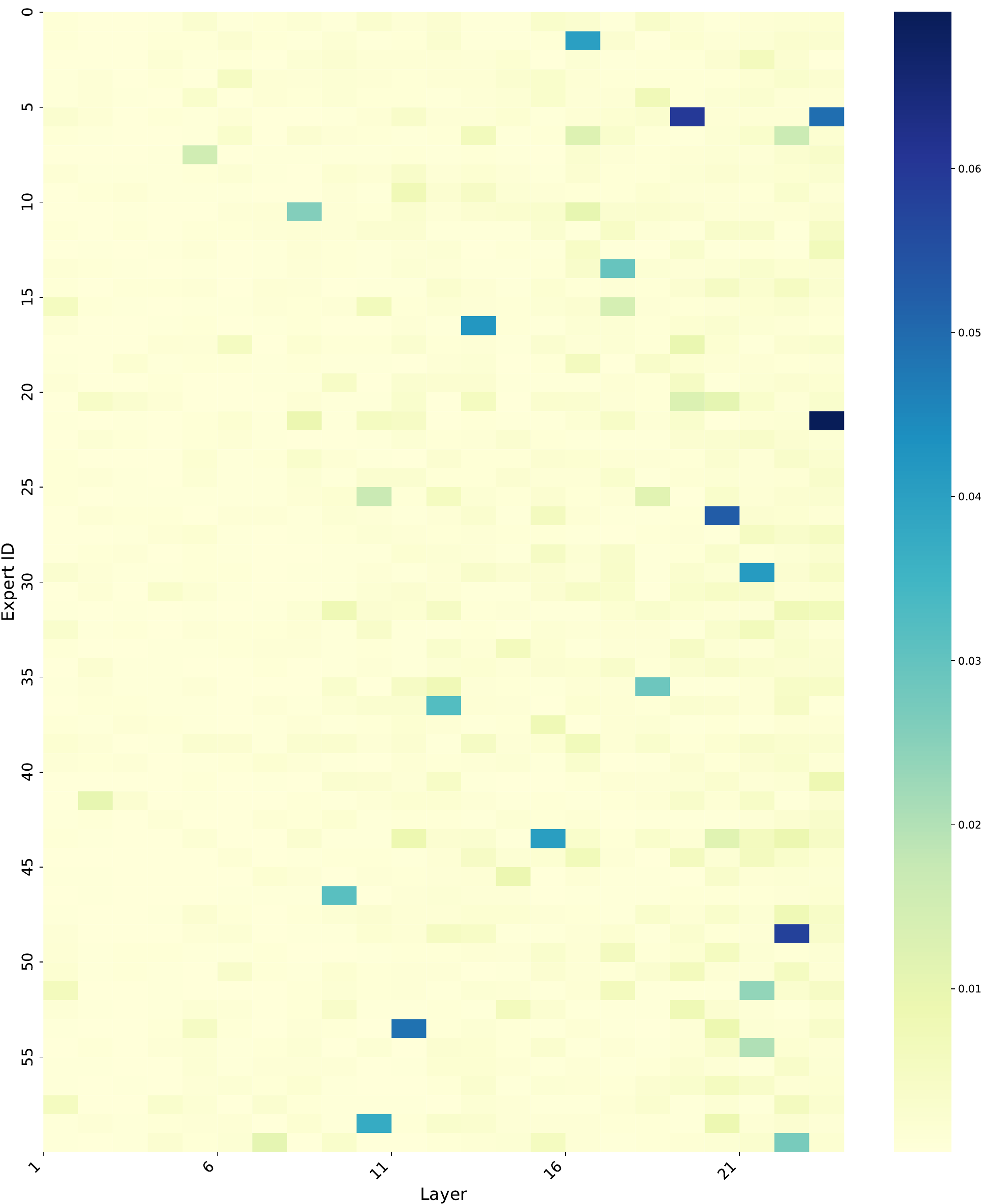}}
\caption{Heatmap of expert-level causal influence across layers for Qwen model. Darker colors indicate stronger causal effects on the model’s output.}
\label{fig:causal_impact_qwen}
\end{figure*}

\begin{figure*}[t]
\centerline{\includegraphics[width=0.9\textwidth]{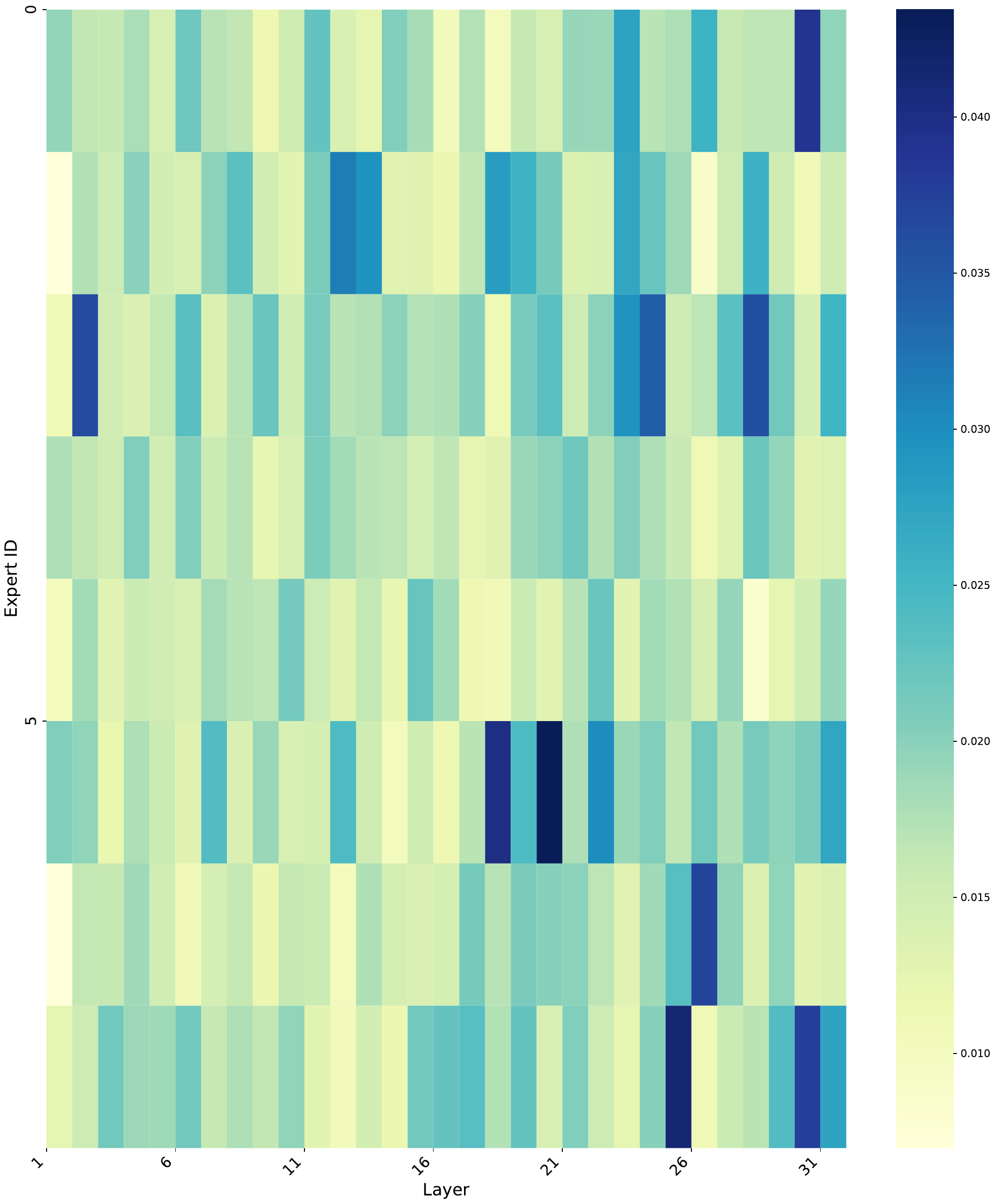}}
\caption{Heatmap of expert-level causal influence across layers for Mixtral model. Darker colors indicate stronger causal effects on the model’s output.}
\label{fig:causal_impact_mixtral}
\end{figure*}

Figures \ref{fig:causal_impact_deepseek}, \ref{fig:causal_impact_qwen} and \ref{fig:causal_impact_mixtral} visualize the causal influence of each expert across all MoE layers for DeepSeek-MoE, Qwen-MoE and Mixtral respectively, where deeper colors indicate stronger causal effects on the model’s output. 
Several observations emerge from the heatmaps across three MoE LLMs: (1) Causal influence is highly uneven across experts and layers.
Only a subset of experts exhibits noticeably strong causal contributions, as indicated by the darkest color blocks. Most experts remain relatively light-colored, suggesting limited impact on the model’s predictions. (2) Causal experts appear intermittently rather than consistently across layers. No single expert dominates all layers. Instead, high-causality activations occur sporadically, implying that different layers rely on different specialized experts to drive key decisions. (3) A small number of experts show concentrated causal spikes.
For example, Expert 2 and Expert 5 display dark blocks in specific layers, revealing moments where their activation plays a decisive role. These localized spikes indicate that certain experts act as “decision drivers” under particular contextual conditions. (4) Upper and middle layers tend to show stronger causal activations.
Compared to early layers, mid-to-late layers contain more high-intensity regions. This suggests that expert activations become more influential as the model approaches deeper semantic representations and final decision formation.
Overall, the heatmap demonstrates that the MoE model relies on a small number of highly decisive experts whose causal impact varies across layers. 
This uneven distribution reinforces the importance of identifying and interpreting these driver experts for deeper understanding of MoE behavior and model interpretability.

\end{document}